\documentclass[11pt,a4paper]{article}
\usepackage[hyperref]{acl2018}
\usepackage{times}
\usepackage{latexsym}

\usepackage{url}

\usepackage{graphicx}
\usepackage{amsmath}
\usepackage{amssymb}
\usepackage{multirow}
\usepackage{booktabs}
\usepackage{epstopdf}

\aclfinalcopy 

\title{Unsupervised Neural Machine Translation with Weight Sharing}

\author{Zhen Yang$^{1,2}$,  Wei Chen$^1$  ,  Feng Wang$^{1,2}$\footnotemark[1], Bo Xu$^1$ \\
  $^1$Institute of Automation, Chinese Academy of Sciences \\
  $^2$University of Chinese Academy of Sciences \\
  {\tt \{yangzhen2014, wei.chen.media, feng.wang, xubo\}@ia.ac.cn}}

\date{}

\begin{document}
\maketitle
\footnotetext[1]{Feng Wang is the corresponding author of this paper}

\begin{abstract}
Unsupervised neural machine translation (NMT) is a recently proposed approach for machine translation which aims to train the model without using any labeled data. The models proposed for unsupervised NMT often use only one shared encoder to map the pairs of sentences from different languages to a shared-latent space, which is weak in keeping the unique and internal characteristics of each language, such as the style, terminology, and sentence structure. To address this issue, we introduce an extension by utilizing two independent encoders but sharing some partial weights which are responsible for extracting high-level representations of the input sentences. Besides, two different generative adversarial networks (GANs), namely the local GAN and global GAN, are proposed to enhance the cross-language translation. With this new approach, we achieve significant improvements on English-German, English-French and Chinese-to-English translation tasks.
\end{abstract}

\section{Introduction}
Neural machine translation \cite{Kalchbrenner:13,sutskever:14,cho:14b,bahdanau:14}, directly applying a single neural network to transform the source sentence into the target sentence, has now reached impressive performance \cite{shen:15,wu2016google,johnson2016google,Gehring2017Convolutional,Vaswani2017Attention}. The NMT typically consists of two sub neural networks. The encoder network reads and encodes the source sentence into a context vector, and the decoder network generates the target sentence iteratively based on the context vector. NMT can be studied in supervised and unsupervised learning settings. In the supervised setting, bilingual corpora is available for training the NMT model. In the unsupervised setting, we only have two independent monolingual corpora with one for each language and there is no bilingual training example to provide alignment information for the two languages. Due to lack of alignment information, the unsupervised NMT is considered more challenging. However, this task is very promising, since the monolingual corpora is usually easy to be collected.

Motivated by recent success in unsupervised cross-lingual embeddings \cite{Artetxe2016Learning,Zhang2017Adversarial,Conneau2017Word}, the models proposed for unsupervised NMT often assume that a pair of sentences from two different languages can be mapped to a same latent representation in a shared-latent space \cite{Lample2017Unsupervised,Artetxe2017Unsupervised}. Following this assumption, \citet{Lample2017Unsupervised} use a single encoder and a single decoder for both the source and target languages. The encoder and decoder, acting as a standard auto-encoder (AE), are trained to reconstruct the inputs. And \citet{Artetxe2017Unsupervised} utilize a shared encoder but two independent decoders. With some good performance, they share a glaring defect, i.e., only one encoder is shared by the source and target languages. Although the shared encoder is vital for mapping sentences from different languages into the shared-latent space, it is weak in keeping the uniqueness and internal characteristics of each language, such as the style, terminology and sentence structure. Since each language has its own characteristics, the source and target languages should be encoded and learned independently. Therefore, we conjecture that the shared encoder may be a factor limiting the potential translation performance.

In order to address this issue, we extend the encoder-shared model, i.e., the model with one shared encoder, by leveraging two independent encoders with each for one language. Similarly, two independent decoders are utilized. For each language, the encoder and its corresponding decoder perform an AE, where the encoder generates the latent representations from the perturbed input sentences and the decoder reconstructs the sentences from the latent representations. To map the latent representations from different languages to a shared-latent space, we propose the weight-sharing constraint to the two AEs. Specifically, we share the weights of the last few layers of two encoders that are responsible for extracting high-level representations of input sentences. Similarly, we share the weights of the first few layers of two decoders. To enforce the shared-latent space, the word embeddings are used as a reinforced encoding component in our encoders. For cross-language translation, we utilize the back-translation following \cite{Lample2017Unsupervised}. Additionally, two different generative adversarial networks (GAN) \cite{Yang2017Improving}, namely the local and global GAN, are proposed to further improve the cross-language translation. We utilize the local GAN to constrain the source and target latent representations to have the same distribution, whereby the encoder tries to fool a local discriminator which is simultaneously trained to distinguish the language of a given latent representation. We apply the global GAN to finetune the corresponding generator, i.e., the composition of the encoder and decoder of the other language, where a global discriminator is leveraged to guide the training of the generator by assessing how far the generated sentence is from the true data distribution \footnote{The code that we utilized to train and evaluate our models can be found at https://github.com/ZhenYangIACAS/unsupervised-NMT}.  In summary, we mainly make the following contributions:

\begin{itemize}
\item We propose the weight-sharing constraint to unsupervised NMT, enabling the model to utilize an independent encoder for each language. To enforce the shared-latent space, we also propose the embedding-reinforced encoders and two different GANs for our model.
\item We conduct extensive experiments on English-German, English-French and Chinese-to-English translation tasks. Experimental results show that the proposed approach consistently achieves great success.
\item Last but not least, we introduce the directional self-attention to model temporal order information for the proposed model. Experimental results reveal that it deserves more efforts for researchers to investigate the temporal order information within self-attention layers of NMT.
\end{itemize}

\section{Related Work}
Several approaches have been proposed to train NMT models without direct parallel corpora. The scenario that has been widely investigated is one where two languages have little parallel data between them but are well connected by one pivot language. The most typical approach in this scenario is to independently translate from the source language to the pivot language and from the pivot language to the target language \cite{saha2016correlational,Cheng2017Joint}. To improve the translation performance, \citet{johnson2016google} propose a multilingual extension of a standard NMT model and they achieve substantial improvement for language pairs without direct parallel training data.

Recently, motivated by the success of cross-lingual embeddings, researchers begin to show interests in exploring the more ambitious scenario where an NMT model is trained from monolingual corpora only. \citet{Lample2017Unsupervised} and \citet{Artetxe2017Unsupervised} simultaneously propose an approach for this scenario, which is based on pre-trained cross lingual embeddings. \citet{Lample2017Unsupervised} utilizes a single encoder and a single decoder for both languages. The entire system is trained to reconstruct its perturbed input. For cross-lingual translation, they incorporate back-translation into the training procedure. Different from \cite{Lample2017Unsupervised}, \citet{Artetxe2017Unsupervised} use two independent decoders with each for one language. The two works mentioned above both use a single shared encoder to guarantee the shared latent space. However, a concomitant defect is that the shared encoder is weak in keeping the uniqueness of each language. Our work also belongs to this more ambitious scenario, and to the best of our knowledge, we are one among the first endeavors to investigate how to train an NMT model with monolingual corpora only.

\begin{figure*}[htb]
	\begin{center}
		\includegraphics[width=9cm]{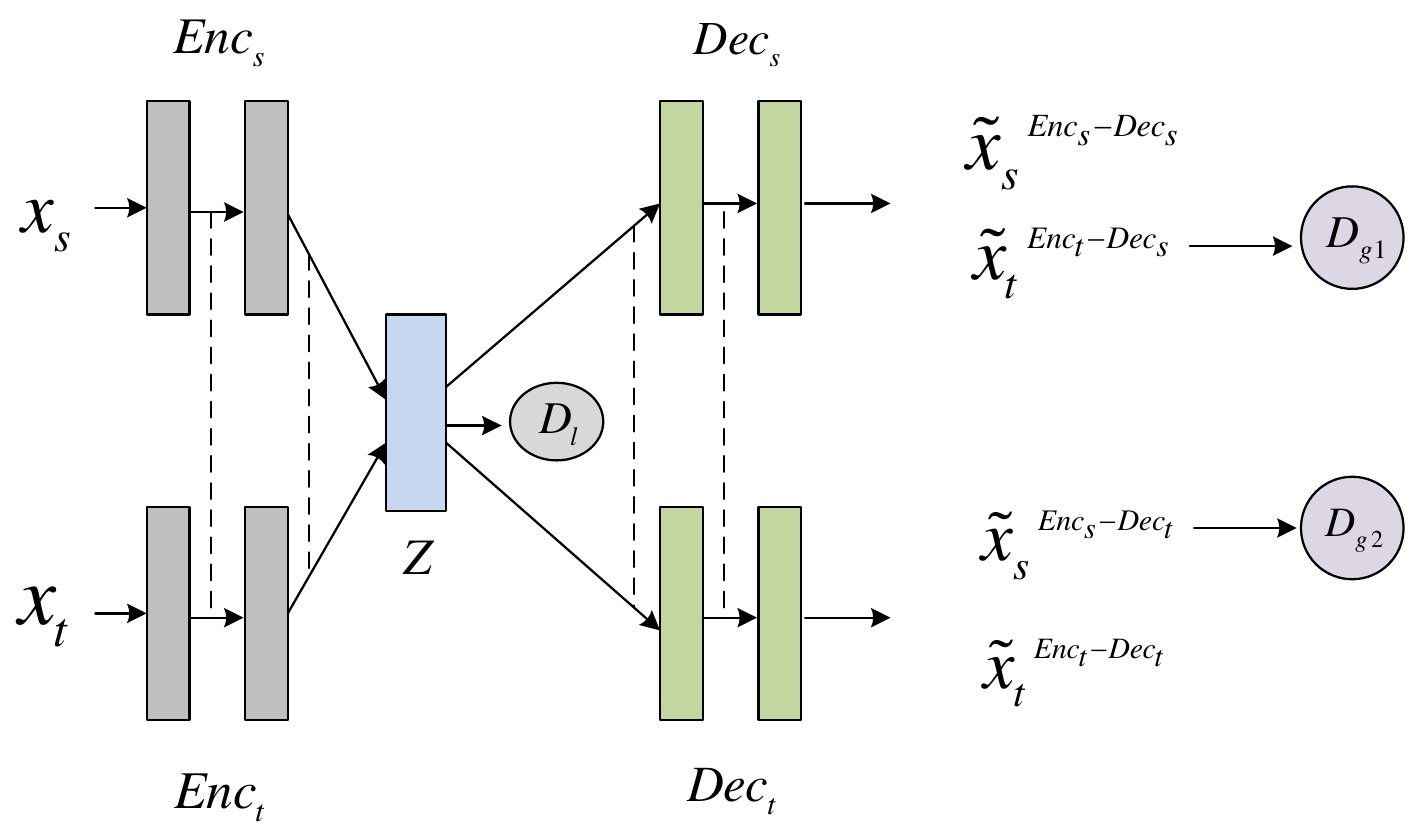}
	\end{center}
	\caption{The architecture of the proposed model. We implement the shared-latent space assumption using a weight sharing constraint where the connection of the last few layers in $Enc_s$ and $Enc_t$ are tied (illustrated with dashed lines) and the connection of the first few layers in $Dec_s$ and $Dec_t$ are tied. $\tilde{x}_s^{Enc_s-Dec_s}$ and $\tilde{x}_t^{Enc_t-Dec_t}$ are self-reconstructed sentences in each language. $\tilde{x}_s^{Enc_s-Dec_t}$ is the translated sentence from source to target and $\tilde{x}_t^{Enc_t-Dec_s}$ is the translation in reversed direction. $D_l$ is utilized to assess whether the hidden representation of the encoder is from the source or target language. $D_{g1}$ and $D_{g2}$ are used to evaluate whether the translated sentences are realistic for each language respectively. $Z$ represents the shared-latent space.}
    \label{model architecture}	
\end{figure*}

\section{The Approach}
\subsection{Model Architecture}
The model architecture, as illustrated in figure \ref{model architecture}, is based on the AE and GAN. It consists of seven sub networks: including two encoders $Enc_{s}$ and $Enc_{t}$, two decoders $Dec_{s}$ and $Dec_{t}$, the local discriminator $D_{l}$, and the global discriminators $D_{g1}$ and $D_{g2}$. For the encoder and decoder, we follow the newly emerged Transformer \cite{Vaswani2017Attention}. Specifically, the encoder is composed of a stack of four identical layers \footnote{The layer number is selected according to our preliminary experiment, which is presented in appendix \ref{sec:appendix for layernum}.}. Each layer consists of a multi-head self-attention and a simple position-wise fully connected feed-forward network. The decoder is also composed of four identical layers. In addition to the two sub-layers in each encoder layer, the decoder inserts a third sub-layer, which performs multi-head attention over the output of the encoder stack. For more details about the multi-head self-attention layer, we refer the reader to \cite{Vaswani2017Attention}. We implement the local discriminator as a multi-layer perceptron and implement the global discriminator based on the convolutional neural network (CNN). Several ways exist to interpret the roles of the sub networks are summarised in table \ref{tab:interpretation_role}. The proposed system has several striking components , which are critical either for the system to be trained in an unsupervised manner or for improving the translation performance.
\begin{table}[htb]
\centering
\scalebox{0.85}{
\begin{tabular}{c|c}
\toprule[2pt]
Networks & Roles \\
\midrule[1pt]
$\{Enc_{s},Dec_{s}\}$ & AE for source language\\
$\{Enc_{t},Dec_{t}\}$ & AE for target language \\
$\{Enc_{s},Dec_{t}\}$ & translation $source \rightarrow target$ \\
$\{Enc_{t},Dec_{s}\}$ & translation $target \rightarrow source$ \\
$\{Enc_{s},D_{l}\}$ & 1st local GAN ($GAN_{l1}$) \\
$\{Enc_{t},D_{l}\}$ & 2nd local GAN ($GAN_{l2}$) \\
$\{Enc_{t},Dec_{s},D_{g1}\}$ & 1st global GAN ($GAN_{g1}$) \\
$\{Enc_{s},Dec_{t},D_{g2}\}$ & 2nd global GAN ($GAN_{g2}$) \\
\bottomrule[2pt]
\end{tabular}}
\caption{Interpretation of the roles for the subnetworks in the proposed system.}
\label{tab:interpretation_role}
\end{table}

\textbf{Directional self-attention}
Compared to recurrent neural network, a disadvantage of the simple self-attention mechanism is that the temporal order information is lost. Although the Transformer applies the positional encoding to the sequence before processed by the self-attention, how to model temporal order information within an attention is still an open question. Following \cite{Shen2017DiSAN}, we build the encoders in our model on the directional self-attention which utilizes the positional masks to encode temporal order information into attention output. More concretely, two positional masks, namely the forward mask $M^f$ and backward mask $M^b$, are calculated as:
\begin{equation} M_{ij}^f = \left\{
\begin{array}{rcl}
0 & &{i < j}\\
-\infty & & otherwise
\end{array}
\right.
\end{equation}
\begin{equation} M_{ij}^b = \left\{
\begin{array}{rcl}
0 & &{i > j}\\
-\infty & & otherwise
\end{array}
\right.
\end{equation}
With the forward mask $M^f$, the later token only makes attention connections to the early tokens in the sequence, and vice versa with the backward mask. Similar to \cite{zhou2016deep,wang2017deep}, we utilize a self-attention network to process the input sequence in forward direction. The output of this layer is taken by an upper self-attention network as input, processed in the reverse direction.

\textbf{Weight sharing} Based on the shared-latent space assumption, we apply the weight sharing constraint to relate the two AEs. Specifically, we share the weights of the last few layers of the $Enc_{s}$ and $Enc_{t}$, which are responsible for extracting high-level representations of the input sentences. Similarly, we also share the first few layers of the $Dec_{s}$ and $Dec_{t}$, which are expected to decode high-level representations that are vital for reconstructing the input sentences. Compared to \cite{cheng2016neural,saha2016correlational} which use the fully shared encoder, we only share partial weights for the encoders and decoders. In the proposed model, the independent weights of the two encoders are expected to learn and encode the hidden features about the internal characteristics of each language, such as the terminology, style, and sentence structure. The shared weights are utilized to map the hidden features extracted by the independent weights to the shared-latent space.

\textbf{Embedding reinforced encoder} We use pre-trained cross-lingual embeddings in the encoders that are kept fixed during training. And the fixed embeddings are used as a reinforced encoding component in our encoder. Formally, given the input sequence embedding vectors $E=\{e_{1},\ldots,e_{t}\}$ and the initial output sequence of the encoder stack $H=\{h_1,\ldots,h_t\}$, we compute $H_r$ as:
\begin{equation}
\label{equa:Hr}
H_{r}=g\odot H+(1-g)\odot E
\end{equation}
where $H_r$ is the final output sequence of the encoder which will be attended by the decoder (In Transformer, $H$ is the final output of the encoder), $g$ is a gate unit and computed as:
\begin{equation}
g=\sigma(W_1 E+ W_2 H + b)
\end{equation}
where $W_1$, $W_2$ and $b$ are trainable parameters and they are shared by the two encoders. The motivation behind is twofold. Firstly, taking the fixed cross-lingual embedding as the other encoding component is helpful to reinforce the shared-latent space. Additionally, from the point of multi-channel encoders \cite{xiong2017multi}, providing encoding components with different levels of composition enables the decoder to take pieces of source sentence at varying composition levels suiting its own linguistic structure.

\subsection{Unsupervised Training}
\label{sec:training procedure}
Based on the architecture proposed above, we train the NMT model with the monolingual corpora only using the following four strategies:

\textbf{Denoising auto-encoding} Firstly, we train the two AEs to reconstruct their inputs respectively. In this form, each encoder should learn to compose the embeddings of its corresponding language and each decoder is expected to learn to decompose this representation into its corresponding language. Nevertheless, without any constraint, the AE quickly learns to merely copy every word one by one, without capturing any internal structure of the language involved. To address this problem, we utilize the same strategy of denoising AE \cite{vincent2008extracting} and add some noise to the input sentences \cite{hill2016learning,Artetxe2017Unsupervised}. To this end, we shuffle the input sentences randomly. Specifically, we apply a random permutation $\varepsilon$ to the input sentence, verifying the condition:
\begin{equation}
|\varepsilon(i)-i|\leq \min(k ([\frac{steps}{s}] + 1), n),    \forall i \in \{1,n\}
\end{equation}
where $n$ is the length of the input sentence, $steps$ is the global steps the model has been updated, $k$ and $s$ are the tunable parameters which can be set by users beforehand. This way, the system needs to learn some useful structure of the involved languages to be able to recover the correct word order. In practice, we set $k=2$ and $s=100000$.

\textbf{Back-translation} In spite of denoising auto-encoding, the training procedure still involves a single language at each time, without considering our final goal of mapping an input sentence from the source/target language to the target/source language. For the cross language training, we utilize the back-translation approach for our unsupervised training procedure. Back-translation has shown its great effectiveness on improving NMT model with monolingual data and has been widely investigated by \cite{sennrich2015improving,Zhang2016Exploiting}.
In our approach, given an input sentence in a given language, we apply the corresponding encoder and the decoder of the other language to translate it to the other language \footnote{Since the quality of the translation shows little effect on the performance of the model \cite{sennrich2015improving}, we simply use greedy decoding for speed.}. By combining the translation with its original sentence, we get a pseudo-parallel corpus which is utilized to train the model to reconstruct the original sentence from its translation.

\textbf{Local GAN} Although the weight sharing constraint is vital for the shared-latent space assumption, it alone does not guarantee that the corresponding sentences in two languages will have the same or similar latent code. To further enforce the shared-latent space, we train a discriminative neural network, referred to as the local discriminator, to classify between the encoding of source sentences and the encoding of target sentences. The local discriminator, implemented as a multi-layer perceptron with two hidden layers of size 256, takes the output of the encoder, i.e., $H_{r}$ calculated as equation \ref{equa:Hr}, as input, and produces a binary prediction about the language of the input sentence. The local discriminator is trained to predict the language by minimizing the following cross-entropy loss:
\begin{equation}
\begin{aligned}
 & L_{D_{l}}(\theta_{D_{l}})=  \\
 & -\mathbb{E}_{x \in x_{s}}[\log p(f=s|Enc_{s}(x))] \\
 & - \mathbb{E}_{x \in x_{t}}[\log p(f=t|Enc_{t}(x))]
 \end{aligned}
\end{equation}
where $\theta_{D_{l}}$ represents the parameters of the local discriminator and $f \in \{s,t\}$. The encoders are trained to fool the local discriminator:
\begin{equation}
\begin{aligned}
& L_{Enc_{s}}(\theta_{Enc_{s}})=   \\
& -\mathbb{E}_{x \in x_{s}}[\log p(f=t|Enc_{s}(x))]
\end{aligned}
\end{equation}

\begin{equation}
\begin{aligned}
& L_{Enc_{t}}(\theta_{Enc_{t}})=   \\
& -\mathbb{E}_{x \in x_{t}}[\log p(f=s|Enc_{t}(x))]
\end{aligned}
\end{equation}
where $\theta_{Enc_{s}}$ and $\theta_{Enc_{t}}$ are the parameters of the two encoders.

\textbf{Global GAN} We apply the global GANs to fine tune the whole model so that the model is able to generate sentences undistinguishable from the true data, i.e., sentences in the training corpus. Different from the local GANs which updates the parameters of the encoders locally, the global GANs are utilized to update the whole parameters of the proposed model, including the parameters of encoders and decoders. The proposed model has two global GANs: $GAN_{g1}$ and $GAN_{g2}$. In $GAN_{g1}$, the $Enc_{t}$ and $Dec_{s}$ act as the generator, which generates the sentence $\tilde{x}_{t}$ \footnote{The $\tilde{x}_t$ is $\tilde{x}_t^{Enc_t-Dec_s}$ in figure \ref{model architecture}. We omit the superscript for simplicity.} from $x_{t}$. The $D_{g1}$, implemented based on CNN, assesses whether the generated sentence $\tilde{x}_t$ is the true target-language sentence or the generated sentence. The global discriminator aims to distinguish among the true sentences and generated sentences, and it is trained to minimize its classification error rate. During training, the $D_{g1}$ feeds back its assessment to finetune the encoder $Enc_t$ and decoder $Dec_s$. Since the machine translation is a sequence generation problem, following \cite{Yang2017Improving}, we leverage policy gradient reinforcement training to back-propagate the assessment. We apply a similar processing to $GAN_{g2}$ (The details about the architecture of the global discriminator and the training procedure of the global GANs can be seen in appendix \ref{sec:appendix for arc of discriminator} and \ref{sec:training procedure of global GAN}).

There are two stages in the proposed unsupervised training. In the first stage, we train the proposed model with denoising auto-encoding, back-translation and the local GANs, until no improvement is achieved on the development set. Specifically, we perform one batch of denoising auto-encoding for the source and target languages, one batch of back-translation for the two languages, and another batch of local GAN for the two languages.  In the second stage, we fine tune the proposed model with the global GANs.

\section{Experiments and Results}
We evaluate the proposed approach on English-German, English-French and Chinese-to-English translation tasks \footnote{The reason that we do not conduct experiments on English-to-Chinese translation is that we do not get public test sets for English-to-Chinese.}. We firstly describe the datasets, pre-processing and model hyper-parameters we used, then we introduce the baseline systems, and finally we present our experimental results.

\subsection{Data Sets and Preprocessing}
In English-German and English-French translation, we make our experiments comparable with previous work by using the datasets from the WMT 2014 and WMT 2016 shared tasks respectively. For Chinese-to-English translation, we use the datasets from LDC, which has been widely utilized by previous works \cite{Tu2017Learning,Zhang2017Prior}.

\textbf{WMT14 English-French} Similar to \cite{Lample2017Unsupervised}, we use the full training set of 36M sentence pairs and we lower-case them and remove sentences longer than 50 words, resulting in a parallel corpus of about 30M pairs of sentences. To guarantee no exact correspondence between the source and target monolingual sets, we build monolingual corpora by selecting English sentences from 15M random pairs, and selecting the French sentences from the complementary set. Sentences are encoded with byte-pair encoding \cite{Sennrich2015Neural}, which has an English vocabulary of about 32000 tokens, and French vocabulary of about 33000 tokens. We report results on $newstest2014$.

\textbf{WMT16 English-German} We follow the same procedure mentioned above to create monolingual training corpora for English-German translation, and we get two monolingual training data of 1.8M sentences each. The two languages share a vocabulary of about 32000 tokens. We report results on $newstest2016$.

\textbf{LDC Chinese-English} For Chinese-to-English translation, our training data consists of 1.6M sentence pairs randomly extracted from LDC corpora \footnote{LDC2002L27, LDC2002T01, LDC2002E18, LDC2003E07, LDC2004T08, LDC2004E12, LDC2005T10}. Since the data set is not big enough, we just build the monolingual data set by randomly shuffling the Chinese and English sentences respectively. In spite of the fact that some correspondence between examples in these two monolingual sets may exist, we never utilize this alignment information in our training procedure (see Section \ref{sec:training procedure}). Both the Chinese and English sentences are encoded with byte-pair encoding. We get an English vocabulary of about 34000 tokens, and Chinese vocabulary of about 38000 tokens. The results are reported on $NIST02$.

Since the proposed system relies on the pre-trained cross-lingual embeddings, we utilize the monolingual corpora described above to train the embeddings for each language independently by using word2vec \cite{mikolov2013distributed}. We then apply the public implementation \footnote{https://github.com/artetxem/vecmap} of the method proposed by \cite{Artetxe2017Learning} to map these embeddings to a shared-latent space \footnote{The configuration we used to run these open-source toolkits can be found in appendix \ref{sec:configuration for toolkits}}.

\subsection{Model Hyper-parameters and Evaluation}
Following the base model in \cite{Vaswani2017Attention}, we set the dimension of word embedding as 512, dropout rate as 0.1 and the head number as 8. We use beam search with a beam size of 4 and length penalty $\alpha=0.6$. The model is implemented in TensorFlow \cite{tensorflow2015-whitepaper} and trained on up to four K80 GPUs synchronously in a multi-GPU setup on a single machine.

For model selection, we stop training when the model achieves no improvement for the tenth evaluation on the development set, which is comprised of 3000 source and target sentences extracted randomly from the monolingual training corpora. Following \cite{Lample2017Unsupervised}, we translate the source sentences to the target language, and then translate the resulting sentences back to the source language. The quality of the model is then evaluated by computing the BLEU score over the original inputs and their reconstructions via this two-step translation process. The performance is finally averaged over two directions, i.e., from source to target and from target to source. BLEU \cite{papineni2002bleu:02} is utilized as the evaluation metric. For Chinese-to-English, we apply the script \emph{mteval-v11b.pl} to evaluate the translation performance. For English-German and English-French, we evaluate the translation performance with the script \emph{multi-belu.pl} \footnote{https://github.com/moses-smt/mosesdecoder/blob/617e8c8/scripts/generic/{multi-bleu.perl;mteval-v11b.pl}}.

\begin{table*}[htb]
			\centering
				\begin{tabular}{c|ccccc}
					\toprule[2pt]
					 & en-de	& de-en	&	en-fr &   fr-en & zh-en\\
					\midrule[1pt]
					Supervised  &   24.07    &   26.99      &  30.50       &   30.21 &  40.02   \\
                    Word-by-word  &   5.85    &  9.34       &   3.60      &  6.80  & 5.09    \\
					\citet{Lample2017Unsupervised}  &  9.64     &    13.33     &     15.05    &    14.31  & - \\
                    \midrule[1pt]
                    \bf{The proposed approach} &    \bf{10.86} & \bf{14.62} & \bf{16.97} & \bf{15.58}& \bf{14.52}  \\
					\bottomrule[2pt]
				\end{tabular}
				\caption{\label{tab:MainResult} The translation performance on English-German, English-French and Chinese-to-English test sets. The results of \cite{Lample2017Unsupervised} are copied directly from their paper. We do not present the results of \cite{Artetxe2017Unsupervised} since we use different training sets.}
\end{table*}

\subsection{Baseline Systems}
\textbf{Word-by-word translation (WBW)} The first baseline we consider is a system that performs word-by-word translations using the inferred bilingual dictionary. Specifically, it translates a sentence word-by-word, replacing each word with its nearest neighbor in the other language.

\textbf{\citet{Lample2017Unsupervised}} The second baseline is a previous work that uses the same training and testing sets with this paper. Their model belongs to the standard attention-based encoder-decoder framework, which implements the encoder using a bidirectional long short term memory network (LSTM) and implements the decoder using a simple forward LSTM. They apply one single encoder and decoder for the source and target languages.

\textbf{Supervised training} We finally consider exactly the same model as ours, but trained using the standard cross-entropy loss on the original parallel sentences. This model can be viewed as an upper bound for the proposed unsupervised model.

\subsection{Results and Analysis}
\subsubsection{Number of weight-sharing layers}
\begin{figure}[htb]
	\begin{center}
		\includegraphics[width=8.1cm]{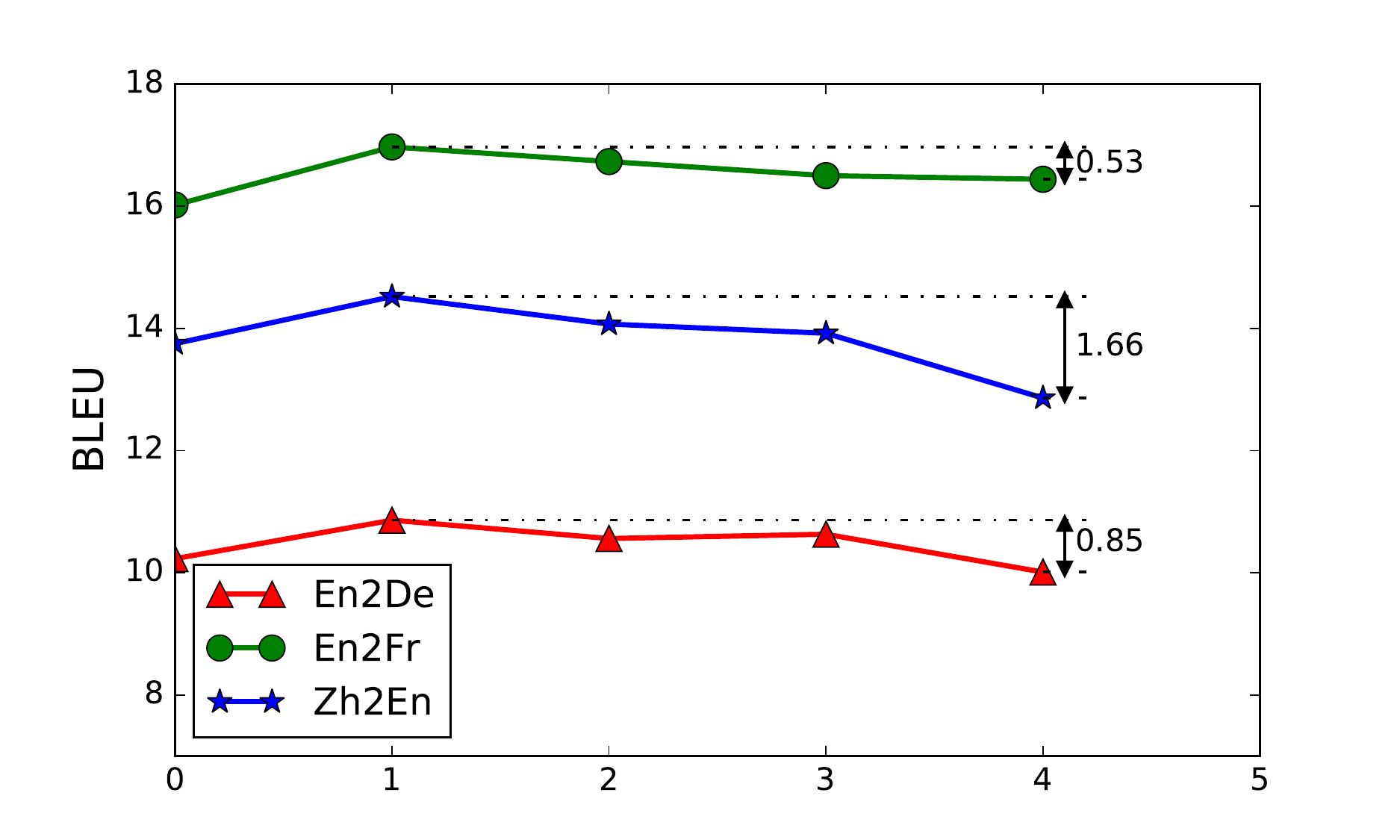}
	\end{center}
	\caption{The effects of the weight-sharing layer number on English-to-German, English-to-French and Chinese-to-English translation tasks.}
    \label{fig:weightSharingNumber}	
\end{figure}

We firstly investigate how the number of weight-sharing layers affects the translation performance. In this experiment, we vary the number of weight-sharing layers in the AEs from 0 to 4. Sharing one layer in AEs means sharing one layer for the encoders and in the meanwhile, sharing one layer for the decoders. The BLEU scores of English-to-German, English-to-French and Chinese-to-English translation tasks are reported in figure \ref{fig:weightSharingNumber}. Each curve corresponds to a different translation task and the x-axis denotes the number of weight-sharing layers for the AEs. We find that the number of weight-sharing layers shows much effect on the translation performance. And the best translation performance is achieved when only one layer is shared in our system. When all of the four layers are shared, i.e., only one shared encoder is utilized, we get poor translation performance in all of the three translation tasks. This verifies our conjecture that the shared encoder is detrimental to the performance of unsupervised NMT especially for the translation tasks on distant language pairs. More concretely, for the related language pair translation, i.e., English-to-French, the encoder-shared model achieves -0.53 BLEU points decline than the best model where only one layer is shared. For the more distant language pair English-to-German, the encoder-shared model achieves more significant decline, i.e., -0.85 BLEU points decline. And for the most distant language pair Chinese-to-English, the decline is as large as -1.66 BLEU points. We explain this as that the more distant the language pair is, the more different characteristics they have. And the shared encoder is weak in keeping the unique characteristic of each language. Additionally, we also notice that using two completely independent encoders, i.e., setting the number of weight-sharing layers as 0, results in poor translation performance too. This confirms our intuition that the shared layers are vital to map the source and target latent representations to a shared-latent space. In the rest of our experiments, we set the number of weight-sharing layer as 1.

\begin{table*}[htb]
			\centering
				\begin{tabular}{c|ccccc}
					\toprule[2pt]
					 & en-de	& de-en	&	en-fr &   fr-en & zh-en\\
					\midrule[1pt]
					Without weight sharing  &  10.23     &  13.84  &     16.02    & 14.82  & 13.75    \\
                    Without embedding-reinforced encoder  &  10.45  &  14.17  & 16.55  & 15.27 & 14.10     \\
					Without directional self-attention  &   10.60 & 14.21  &  16.82   &  15.30  &  14.29     \\
                    Without local GANs &  10.51     &   14.35      &   16.40          &  15.07    &  14.12   \\
                    Without Global GANs  &    10.34   &  14.05       &  16.19         &  15.21    &  14.09   \\
                    \textbf{Full model}&\textbf{10.86}&\textbf{14.62}&\textbf{16.97}&\textbf{15.58}&\textbf{14.52}    \\
					\bottomrule[2pt]
				\end{tabular}
				\caption{\label{tab:Ablation} Ablation study on English-German, English-French and Chinese-to-English translation tasks. Without weight sharing means no layers are shared in the two AEs.}
\end{table*}

\subsubsection{Translation results}
Table \ref{tab:MainResult} shows the BLEU scores on English-German, English-French and English-to-Chinese test sets. As it can be seen, the proposed approach obtains significant improvements than the word-by-word baseline system, with at least +5.01 BLEU points in English-to-German translation and up to +13.37 BLEU points in English-to-French translation. This shows that the proposed model only trained with monolingual data effectively learns to use the context information and the internal structure of each language. Compared to the work of \cite{Lample2017Unsupervised}, our model also achieves up to +1.92 BLEU points improvement on English-to-French translation task. We believe that the unsupervised NMT is very promising. However, there is still a large room for improvement compared to the supervised upper bound. The gap between the supervised and unsupervised model is as large as 12.3-25.5 BLEU points depending on the language pair and translation direction.

\subsubsection{Ablation study}
To understand the importance of different components of the proposed system, we perform an ablation study by training multiple versions of our model with some missing components: the local GANs, the global GANs, the directional self-attention, the weight-sharing, the embedding-reinforced encoders, etc. Results are reported in table \ref{tab:Ablation}. We do not test the the importance of the auto-encoding, back-translation and the pre-trained embeddings because they have been widely tested in \cite{Lample2017Unsupervised,Artetxe2017Unsupervised}. Table \ref{tab:Ablation} shows that the best performance is obtained with the simultaneous use of all the tested elements. The most critical component is the weight-sharing constraint, which is vital to map sentences of different languages to the shared-latent space. The embedding-reinforced encoder also brings some improvement on all of the translation tasks. When we remove the directional self-attention, we get up to -0.3 BLEU points decline. This indicates that it deserves more efforts to investigate the temporal order information in self-attention mechanism. The GANs also significantly improve the translation performance of our system. Specifically, the global GANs achieve improvement up to +0.78 BLEU points on English-to-French translation and the local GANs also obtain improvement up to +0.57 BLEU points on English-to-French translation. This reveals that the proposed model benefits a lot from the cross-domain loss defined by GANs.

\section{Conclusion and Future work}
The models proposed recently for unsupervised NMT use a single encoder to map sentences from different languages to a shared-latent space. We conjecture that the shared encoder is problematic for keeping the unique and inherent characteristic of each language. In this paper, we propose the weight-sharing constraint in unsupervised NMT to address this issue. To enhance the cross-language translation performance, we also propose the embedding-reinforced encoders, local GAN and global GAN into the proposed system. Additionally, the directional self-attention is introduced to model the temporal order information for our system.

We test the proposed model on English-German, English-French and Chinese-to-English translation tasks. The experimental results reveal that our approach achieves significant improvement and verify our conjecture that the shared encoder is really a bottleneck for improving the unsupervised NMT. The ablation study shows that each component of our system achieves some improvement for the final translation performance.

Unsupervised NMT opens exciting opportunities for the future research. However, there is still a large room for improvement compared to the supervised NMT. In the future, we would like to investigate how to utilize the monolingual data more effectively, such as incorporating the language model and syntactic information into unsupervised NMT. Besides, we decide to make more efforts to explore how to reinforce the temporal order information for the proposed model.

\section*{Acknowledgements}
This work is supported by the National Key Research and Development Program of China under Grant No. 2017YFB1002102, and Beijing Engineering Research Center under Grant No. Z171100002217015. We would like to thank Xu Shuang for her preparing data used in this work. Additionally, we also want to thank Jiaming Xu, Suncong Zheng and Wenfu Wang for their invaluable discussions on this work.

\bibliography{acl2018}
\bibliographystyle{acl_natbib}

\appendix
\section{Experiments on the layer number for encoders and decoders}
\label{sec:appendix for layernum}
To determine the number of layers for encoders and decoders in our system beforehand, we conduct experiments on English-German translation tasks to test how the amount of layers in encoders and decoders affects the translation performance. We vary the number of layers from 2 to 6 and the results are reported in table \ref{tab:layer num}. We can find that the translation performance achieves substantial improvement with the layer number increasing from 2 to 4. However, with layer number set larger than 4, we get little improvement. To make a trade-off between the translation performance and the computation complexity, we set the layer number as 4 for our encoders and decoders.
\begin{table}[htb]
			\centering
				\begin{tabular}{c|cc}
					\toprule[2pt]
					layer num & en-de	& de-en	 \\
					\midrule[1pt]
                    2  &   11.57    &   14.01      \\
					3  &   12.43    &   14.99      \\
                    4  &   12.86    &   15.62      \\
                    5  &   12.91    &   15.83      \\
                    6  &   12.95    &   15.79      \\
					\bottomrule[2pt]
				\end{tabular}
				\caption{\label{tab:layer num} The experiments on the number of layers for encoders and decoders.}
\end{table}

\section{The architecture of the global discriminator}
\label{sec:appendix for arc of discriminator}
The global discriminator is applied to classify the generated sentences as source language, target language or generated sentences. Following \cite{Yang2017Improving}, we implement the global discriminator based on CNN. Since sentences generated by the generator (the composition of the encoder and decoder) have variable lengths, the CNN padding is used to transform the sentences to sequences with fixed length $T$, which is the maximum length set for the output of the generator. Given the generated sequences $x_1, \ldots, x_T$, we build the matrix $X_{1:T}$ as:
\begin{equation}X_{1:T}=x_1;x_2;\ldots;x_T\end{equation}
where $x_t \in R^k$ is the $k$-dimensional word embedding and the semicolon is the concatenation operator. For the matrix $X_{1:T}$, a kernel ${w_j}\in R^{l \times k}$ applies a convolutional operation to a window size of $l$ words to produce a series of feature maps:
\begin{equation}\label{equa:BN}c_{ji}=\rho(BN(w_j \otimes X_{i:i+l-1}+b))\end{equation}
where $\otimes$ operator is the summation of element-wise production and $b$ is a bias term. $\rho$ is a non-linear activation function which is implemented as ReLu in this paper. To get the final feature with respect to kernel $w_j$, a max-over-time pooling operation is leveraged over the feature maps:
\begin{equation}\label{equa:c}\widetilde{c}_j=max\{c_{j1},\ldots,c_{j{T-l+1}}\}\end{equation}
We use various numbers of kernels with different window sizes to extract different features, which are then concatenated to form the final sentence representation $x_c$. Finally, we pass $x_c$ through a fully connected layer and a softmax layer to generate the probability $p(f_g|x_1,\ldots,x_T)$ as:
\begin{equation}p(f_g|x_1,\ldots,x_T)=softmax(V*x_c)\end{equation}
where $V$ is the transformation matrix and $f_g\in \{true, generated \}$.

\section{The training procedure of the global GAN}
\label{sec:training procedure of global GAN}
We apply the global GANs to finetune the whole model. Here, we provide detailed strategies for training the global GANs. Firstly, we generate the machine-generated source language sentences by using $Enc_t$ and $Enc_s$ to decode the monolingual data in target language. Similarly, we get the generated sentences in target language with $Enc_s$ and $Dec_t$ by decoding source language monolingual data. We simply use the greedy sampling method instead of the beam search method for decoding. Next, we pre-train $D_{g1}$ on the combination of true monolingual data and the generated data in the source language. Similarly, we also pre-train $D_{g2}$ on the combination of true monolingual data and the generated data in the target language. Finally, we jointly train the generators and discriminators. The generators are trained with policy gradient training methods. For the details about the policy gradient training, we refer the reader to \cite{Yang2017Improving}.

\section{The configurations for the open-source toolkits}
\label{sec:configuration for toolkits}
We train the word embedding use the following script:

\emph{./word2vec -train text -output embedding.txt -cbow 0 -size 512 -window 10 -negative 10 -hs 0 -sample 1e- -threads 50 -binary 0 -min-count 5 -iter 10}

After we get the embeddings for both the source and target languages, we use the open-source VecMap \footnote{https://github.com/artetxem/vecmap} to map these embeddings to a shared-latent space with the following scripts:

\emph{python3 normalize\_embeddings.py unit center -i s\_embedding.txt -o s\_embedding.normalized.txt}

\emph{python3 normalize\_embeddings.py unit center -i t\_embedding.txt -o t\_embedding.normalized.txt}

\emph{python3 map\_embeddings.py --orthogonal s\_embedding.normalized.txt t\_embedding.normalized.txt s\_embedding.mapped.txt t\_embedding.mapped.txt --numerals --self\_learning -v}

\end{document}